\documentclass[conference]{IEEEtran}
\usepackage{epsfig,endnotes}
\usepackage{url}
\usepackage{subfigure}
\usepackage{rotating}
\usepackage{float}
\usepackage{alltt}
\usepackage{color,colortbl}
\definecolor{Gray}{gray}{0.9}
\definecolor{LightCyan}{rgb}{0.88,1,1}
\usepackage{amssymb,amsmath}
\usepackage{xspace}
\usepackage{booktabs}
\usepackage{enumitem}
\usepackage{tablefootnote}
\usepackage{listings}
\usepackage{natbib}
\newcommand{\Tool}{{Rudra}\xspace}
\newcommand{\Base}{{Rudra-base}\xspace}
\newcommand{\Adv}{{Rudra-adv}\xspace}
\newcommand{\Advstar}{{Rudra-adv\ts{$\ast$}}\xspace}

\newcommand{\minibatch}{{mini-batch}\xspace}
\newcommand{\CIFAR}{{\tt{CIFAR10}}\xspace}
\newcommand{\imagenet}{{\tt{ImageNet}}\xspace}
\newcommand{\SML}{$(\sigma,\mu,\lambda)$\xspace}
\newcommand{\pullWeights}{{\tt{pullWeights}}\xspace}
\newcommand{\pushGradient}{{\tt{pushGradient}}\xspace}
\newcommand{\getMinibatch}{{\tt{getMinibatch}}\xspace}
\newcommand{\calcGradient}{{\tt{calcGradient}}\xspace}
\newcommand{\applyUpdate}{{\tt{applyUpdate}}\xspace}
\newcommand{\sumGradients}{{\tt{sumGradients}}\xspace}
\newcommand{\cfga}{{\textit{base-hardsync}}\xspace}
\newcommand{\cfgb}{{\textit{base-softsync}}\xspace}
\newcommand{\cfgc}{{\textit{adv-softsync}}\xspace}
\newcommand{\cfgd}{{\textit{adv\ts{$\ast$}-softsync}}\xspace}
\newcommand{\ts}{\textsuperscript}
\newcommand\T{\rule{0pt}{2.6ex}}       
\newcommand\B{\rule[-1.2ex]{0pt}{0pt}} 
\makeatletter
\renewcommand{\paragraph}{%
  \@startsection{paragraph}{4}%
  {\z@}{0.5ex \@plus 0.2ex \@minus 0.1ex}{-1em}%
  {\normalfont\normalsize\bfseries}%
}
\usepackage[binary-units=true]{siunitx}

\begin{document}
%
%
\title{Model Accuracy and Runtime Tradeoff in Distributed Deep Learning: A Systematic Study}
\author{\IEEEauthorblockN{Suyog Gupta$^{1,*}$, Wei Zhang$^{1,*}$, Fei Wang$^2$}
\IEEEauthorblockA{$^1$IBM T. J. Watson Research Center. Yorktown Heights. NY.\quad $^*$Equal Contribution.\\
$^2$Department of Healthcare Policy and Research, Weill Cornell Medical College. New York City. NY}
\vspace{-2em}}

\maketitle
\begin{abstract}
Deep learning with a large number of parameters requires distributed training, where model accuracy and runtime are two important factors to be considered.
However, there has been no systematic study of the tradeoff between these two factors during the model training process. 
This paper presents Rudra, a parameter server based distributed computing framework tuned for training large-scale deep neural networks. Using variants of the asynchronous stochastic gradient descent algorithm we study the impact of synchronization protocol, stale gradient updates, minibatch size, learning rates, and number of learners on runtime performance and model accuracy. We introduce a new learning rate modulation strategy to counter the effect of stale gradients and propose a new synchronization protocol that can effectively bound the staleness in gradients, improve runtime performance and achieve good model accuracy. Our empirical investigation reveals a principled approach for distributed training of neural networks: the mini-batch size per learner should be reduced as more learners are added to the system to preserve the model accuracy. We validate this approach using commonly-used image classification benchmarks: CIFAR10 and ImageNet.
\end{abstract}

\section{Introduction}
Deep neural network based models have achieved unparalleled accuracy in cognitive tasks such as speech recognition, object detection, and natural language processing~\cite{lecun2015deep}. 
For certain image classification benchmarks, deep neural networks have been touted to even surpass human-level performance~\cite{ioffe2015batch,he2015delving}. 
Such accomplishments are made possible by the ability to perform fast, supervised training  of complex neural network architectures using large  quantities  of labeled data. 
Training a deep neural network translates into solving a non-convex optimization problem in a very high dimensional space, and in the absence of a solid theoretical framework to solve such problems, practitioners are forced to rely on trial-and-error empirical observations to design heuristics that help obtain a well-trained model\cite{bengio2012practical}. 
Naturally, fast training of deep neural network models can enable rapid evaluation of different network architectures and facilitate a more thorough hyper-parameter optimization for these models. Recent years have seen a resurgence of interest in deploying large-scale  computing infrastructure designed specifically for training deep  neural  networks. 
Some notable efforts in this direction include distributed computing infrastructure  using  thousands  of  CPU cores \cite{adam, distbelief}, high-end graphics processors (GPUs)\cite{krizhevsky2012imagenet}, or a combination of CPUs and GPUs \cite{coates2013deep}.

The large-scale deep learning problem can hence be viewed as a confluence of elements from machine learning (ML) and high-performance computing (HPC). 
Much of the work in the ML community is focused on non-convex optimization, model selection, and hyper-parameter tuning to improve the neural network's performance (measured as classification accuracy) while working under the constraints of the computational resources available in a single computing node (CPU with or without GPU acceleration).
From a HPC perspective, prior work has addressed, to some extent, the problem of accelerating the neural network training by mapping the asynchronous version of mini-batch stochastic gradient descent (SGD) algorithm onto multiple computing nodes. Contrary to the popular belief that asynchrony necessarily improves model accuracy, we find that adopting the approach of scale-out deep learning using asynchronous-SGD, gives rise to complex interdependencies between the training algorithm's hyperparameters and the distributed implementation's design choices (synchronization protocol, number of learners), ultimately impacting the neural network's accuracy and the overall system's runtime performance. 

In this paper we present \emph{\Tool}, a parameter server based deep learning framework to study these interdependencies and undertake an empirical evaluation with public image classification benchmarks. Our key contributions are:
\begin{enumerate}
  \item A systematic technique (vector clock) for quantifying the staleness of gradient descent parameter updates.
  \item An investigation of the impact of the interdependence of training algorithm's hyperparameters (\minibatch size, learning rate (gradient descent step size)) and distributed implementation's parameters (gradient staleness, number of learners) on the neural network's classification accuracy and training time.
  \item A new learning rate tuning strategy that reduces the effect of stale parameter updates.
    \item A new synchronization protocol to reduce network bandwidth overheads while achieving good classification accuracy and runtime performance.
  \item An observation that to maintain a given level of model accuracy, it is necessary to reduce the \minibatch size as the number of learners is increased.
  This suggests a hard limit on the amount of parallelism that can be exploited in training a given model.
\end{enumerate}

\section{Background}
A neural network computes a parametric, non-linear transformation $f_\theta:X \mapsto Y$, where $\theta$ represents a set of adjustable parameters (or weights). 
In a supervised learning context (such as image classification), $X$ is the input image and $Y$ corresponds to the label assigned to the image. 
A deep neural network organizes the parameters $\theta$ into multiple \emph{layers}, each of which consists of a linear transformation followed by a non-linear function such as sigmoid, tanh, etc. 
In a feed-forward deep neural network, the layers are arranged hierarchically such that the output of the layer $l-1$ feeds into the input of layer $l$. 
The terminal layer generates the network's output $\hat{Y} = f_{\theta}(X)$, corresponding to the input $X$.

A neural network training algorithm seeks to find a set of parameters $\theta^{\ast}$ that minimizes the discrepancy between $\tilde{Y}$ and the ground truth $Y$. 
This is usually accomplished by defining a differentiable cost function $C(\hat{Y},Y)$ and iteratively updating each of the model parameters using some variant of the gradient descent algorithm:{\small
\begin{subequations}\label{eq:gd}
\begin{align}
E_m &= \frac{1}{m}\sum\nolimits_{s=1}^{m}C \left(\hat{Y_s},Y_s\right), \label{eq:gd1} \\
\nabla\theta^{(k)}(t) &= \left({\partial E_m}/ {\partial \theta^{(k)}}\right)(t), \label{eq:gd2} \\
\theta^{(k)}(t+1) &= \theta^{(k)}(t) - \alpha(t)\nabla\theta^{(k)}(t) \label{eq:gd3}
\end{align}
\end{subequations}}
where $\theta^{(k)}(t)$ represents the $k^{th}$ parameter at iteration $t$, $\alpha$ is the step size (also known as the learning rate) and $m$ is the batch size. 
The batch gradient descent algorithm sets $m$ to be equal to the total number of training examples $N$.
Due to the large amount of training data, deep neural networks are typically trained using the Stochastic Gradient Descent (SGD), where the parameters are updated with a randomly selected training example $(X_s,Y_s)$. 
The performance of SGD can be improved by computing the gradients using a \emph{\minibatch} containing $m = \mu \ll N$ training examples.

Deep neural networks are generally considered hard to train \cite{bengio2012practical,glorot2010understanding,sutskever2013importance} and the trained model's generalization error depends strongly on hyperparameters such as the initializations, learning rates, \minibatch size, network architecture, etc. 
In addition, neural networks are prone to overfit the data. Regularization methods (e.g., weight decay and dropout) \cite{krizhevsky2012imagenet} applied during training have been shown to combat overfitting and reduce the generalization error.\\
\underline{\textit{Scale-out deep learning}}: A typical implementation of distributed training of deep neural networks consists of a master~(parameter server) that orchestrates the work among one or more slaves~(learners).
Each learner does the followings:
\begin{enumerate}
\item \getMinibatch : Select randomly a \minibatch of examples from the training data.
\item \pullWeights : Request the parameter server for the current set of weights/parameters.
\item \calcGradient : Compute gradients based on the training error for the current \minibatch (equation~\ref{eq:gd2}).
\item \pushGradient : Send the computed gradients to the parameter server
\end{enumerate}

The parameter server maintains a global view of the model weights and performs the following functions:\vspace{-.5em}
\begin{enumerate}
\item \sumGradients : Receive and accumulate the gradients from the learners.
\item \applyUpdate : Multiply the accumulated gradient by the learning rate and update the weights (equation~\ref{eq:gd3})
\end{enumerate}

Learners exploit \emph{data parallelism} by each maintaining a copy of the entire model, and training independently over a unique \minibatch.
The \emph{model parallelism} approach augments this framework by splitting the neural network model across multiple learners. 
With model parallelism, each learner trains only a portion of the network; edges that cross learner boundaries must be synchronized before gradients can be computed for the entire model.

Several different synchronization strategies are possible.
The most commonly used one is the asynchronous protocol, in which the learners work completely independently of each other and the parameter server. 
Section~\ref{sec:design} will discuss three different synchronization strategies, each associated with a unique tradeoff between model accuracy and runtime.

\section{Design and Implementation}
\label{sec:design}
\subsection{\label{sec:term_definitions}Terminology}
Throughout the paper, we use the following definitions:\vspace{-.5em}
\begin{itemize}
\item Parameter Server: a server that holds the model weights. 
\cite{parameterserver} describes a typical parameter server using a distributed key-value store to synchronize state between processes.
The parameter server collects gradients from learners and updates the weights accordingly.
\item Learner: A computing process that can calculate weight updates (gradients).
\item $\mu$: mini-batch size. 

\item $\alpha$: learning rate.

\item $\lambda$: number of learners.

\item Epoch: a pass through the entire training dataset.

\item Timestamp: we use a scalar clock~\cite{logicaltime} to represent weights timestamp $ts_{i}$, starting from $i=0$.
Each weight update increments the timestamp by 1.
The timestamp of a gradient is the same as the timestamp of the weight used to compute the gradient.

\item $\sigma$: staleness of the gradient.  A gradient with timestamp $ts_i$ is pushed to the parameter server with current weight timestamp $ts_j$, where $ts_j \geq ts_i$.
We define the staleness of this gradient $\sigma$ as $j-i$.

\item $\langle\sigma\rangle$, average staleness of gradients.
The timestamps of the set of $n$ gradients that triggers the advancement of weights timestamp from $ts_{i-1}$ to $ts_{i}$ form a vector clock~\cite{vectorclock} $\langle ts_{i_{1}}, ts_{i_{2}}, ..., ts_{i_{n}}\rangle$, where $max\{i_{1}, i_{2}, ..., i_{n}\} < i$.
The average staleness of gradients $\langle\sigma\rangle$ is defined as:{\small
\begin{equation}
\langle\sigma\rangle =  (i-1) - mean(i_{1}, i_{2}, ..., i_{n})\label{eqn:avg_staleness}
\end{equation}}

\item Hardsync protocol: To advance weights timestamp from $ts_{i}$ to $ts_{i+1}$, each learner calculates exactly one \minibatch and sends its gradient $\nabla\theta_{l}$ to the parameter server.
The parameter server averages the gradients and updates the weights according to Equation~\eqref{eqn:update_hard}, then broadcasts the new weights to all learners.
Staleness in the hardsync protocol is always zero.{\small
\begin{equation}
	\begin{aligned}
		\nabla \theta^{(k)}(i) &= \frac{1}{\lambda}{\sum\nolimits_{l=1}^{\lambda} \nabla\theta_{l}^{(k)}}\\
		\theta^{(k)}(i+1) &= \theta^{(k)}(i) - \alpha\nabla \theta^{(k)}(i)
	\end{aligned}
 \label{eqn:update_hard}
\end{equation}}\vspace{-.5em}

\item Async protocol: Each learner calculates the gradients and asynchronously pushes/pulls the gradients/weights to/from parameter server. The Async weight update rule is given by:{\small
\begin{equation}
\begin{aligned}
\nabla \theta^{(k)}(i) &= \nabla\theta_{l}^{(k)}, L_{l} \in {L_{1}, ..., L_{\lambda}} \\
\theta^{(k)}(i+1) &= \theta^{(k)}(i) - \alpha\nabla \theta^{(k)}(i)
\end{aligned} 
\label{eqn:update_async}
\end{equation}}
Gradient staleness may be hard to control due to the asynchrony in the system.
\cite{distbelief} describe \emph{Downpour SGD}, an implementation of the Async protocol for a commodity scale-out system in which the staleness can be as large as hundreds.
 
\item $n$-softsync protocol: Each learner pulls the weights from the parameter server, calculates the gradients and pushes the gradients to the parameter server.
The parameter server updates the weights after collecting at least $c =  \lfloor(\lambda / n) \rfloor$ gradients.
The splitting parameter $n$ can vary from 1 to $\lambda$.
The $n$-softsync weight update rule is given by:{\small
\begin{equation}
	\begin{aligned}
		c &= \lfloor(\lambda / n) \rfloor \\
		\nabla \theta^{(k)}(i) &= \frac{1}{c}{\sum\nolimits_{l=1}^c  \nabla\theta_{l}^{(k)}}, L_{j} \in {L_{1}, ..., L_{\lambda}} \\
		\theta^{(k)}(i+1) &= \theta^{(k)}(i) - \alpha\nabla \theta^{(k)}(i)
 \label{eqn:update_soft}
 \end{aligned}
\end{equation}}

In Section~\ref{sec:eval:staleness} we will show that in a homogeneous cluster where each learner proceeds at roughly the same speed, the staleness of the model can be empirically bounded at $2n$. \textbf{Note that when $n$ is equal to $\lambda$, the weight update rule at the parameter server is exactly the same as in Async protocol.}

\end{itemize}

\subsection{\Base System Architecture}
\label{sec:base_arch}
\begin{figure}[t]
\centering
\includegraphics[width=0.65\columnwidth]{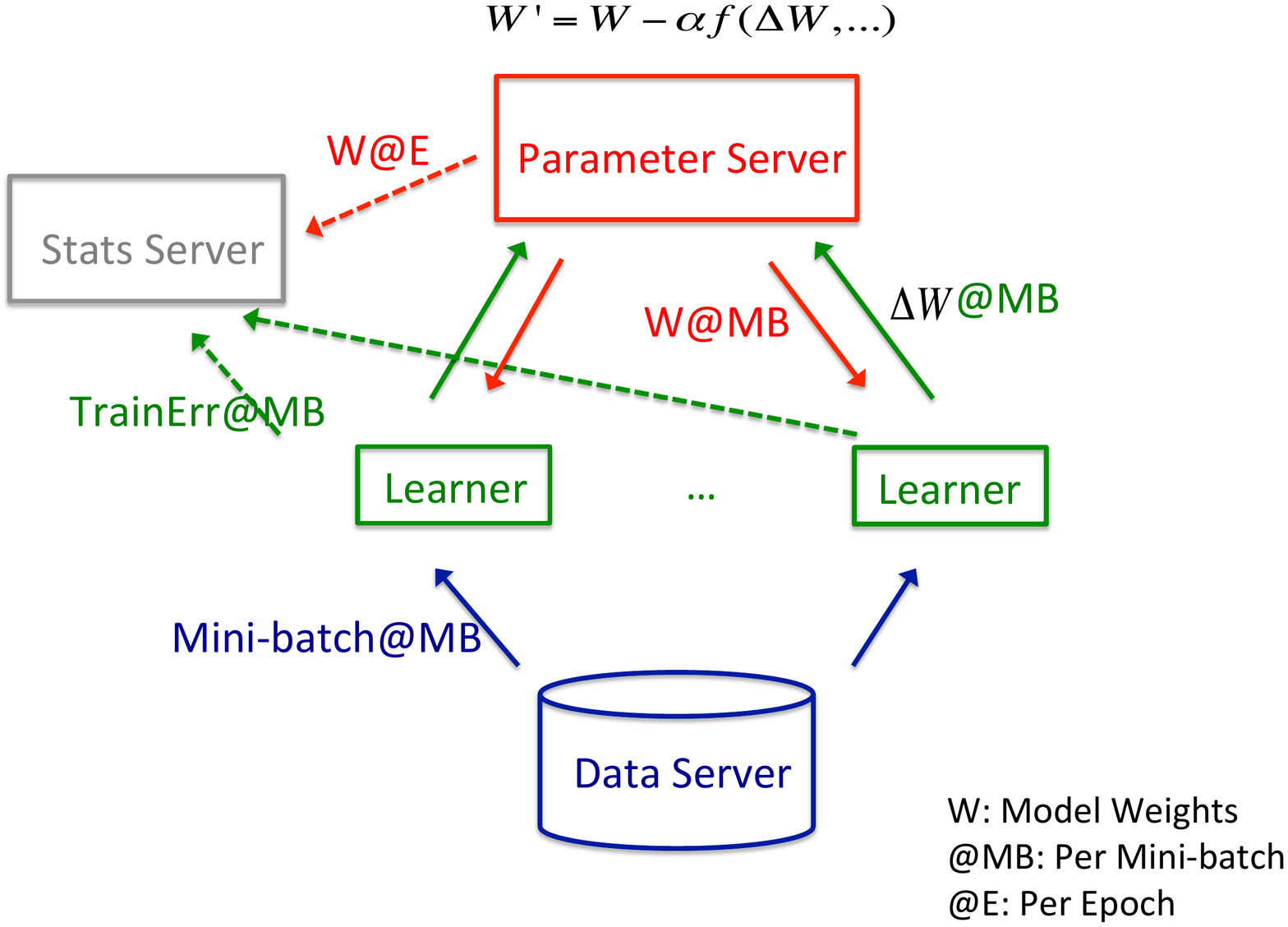}
\vspace{-1mm}
\caption{\label{fig:base_arch} \Base architecture }
\vspace{-4mm}
\end{figure}
Figure~\ref{fig:base_arch} illustrates the parameter server design that we use to study the interplay of hyperparameter tuning and system scale-out factor. This system implements both hardsync and n-softsync protocols. The arrows between each entity represent a (group of) MPI message(s), except the communication between \textit{Learner} and \textit{Data Server}, which is achieved by a global file system. We describe each entity's role and its implementation below.\\
{\bf \underline{\textit{Learner}}} is a single-process multithreaded SGD solver. 
Before training each mini-batch, a learner pulls the weights and the corresponding timestamp from the parameter server. A learner reduces the \pullWeights traffic by first inquiring the timestamp from the parameter server: if the timestamp is as old as the local weights', then this learner does not pull the weights. After training the mini-batch, learner sends the gradients along with gradients' timestamp to parameter server.  The size of pull and push messages is the same as the model size plus the size of scalar timestamp equal to one.  \\
{\bf \underline{\textit{Data Server}}} is hosted on IBM GPFS, a global file system.
Each learner has an I/O thread, which prefetches the \minibatch via random sampling prior to training.
Prefetching is completely overlapped with the computation.  \\
{\bf \underline{\textit{Parameter Server}}} is a multithreaded process, that accumulates gradients from each learner and applies update rules according to Equations~(\ref{eqn:update_hard}--\ref{eqn:update_soft}).
In this study, we implemented hardsync protocol and $n$-softsync protocol.
Learning rate is configured differently in either protocol.
In hardsync protocol, the learning rate is multiplied by a factor $\sqrt{\lambda\mu / B}$, where $B$ is the batch size of the reference model. In the $n$-softsync protocol, the learning rate is multiplied by the reciprocal of staleness. We demonstrate in Section~\ref{sec:eval:staleness} that this treatment of learning rate in $n$-softsync can significantly improve the model accuracy. Parameter server records the vector clock of each weight update to keep track of the the average staleness. When a specified number of epochs are trained, parameter server shuts down each learner.\\ 
{\bf \underline{\textit{Statistics Server}}} is a multithreaded process that receives the training error from each learner and receives the model from the parameter server at the end of each epoch and tests the model. It monitors the model training quality.  

This architecture is non-blocking everywhere except for pushing up gradients and pushing down weights, which are blocking MPI calls (e.g. \texttt{MPI\_Send}).
Parameter server handles each incoming message one by one (the message handling itself is multithreaded).
In this way, we can precisely control how each learner's gradients are received and handled by the parameter server.
The purpose of \Base is to control the noise of the system, so that we can study the interplay of scale-out factor and the hyperparameter selection.
For a moderately-sized dataset like CIFAR-10, \Base shows good scale-out factor (see Section~\ref{sec:tradeoffcurves}).

\subsection{\Adv and \Advstar System Architecture}

To achieve high classification accuracy, the required model size may be quite large (e.g. hundreds of MBs).
In many cases, to achieve best possible model accuracy, \minibatch size $\mu$ must be small, as we will demonstrate in Section~\ref{sec:tradeoffcurves}.
In order to meet these requirements with acceptable performance, we implemented \Adv and \Advstar.
\begin{figure}[t]
\centering
\subfigure[\Adv architecture]{\includegraphics[width=0.4\columnwidth]{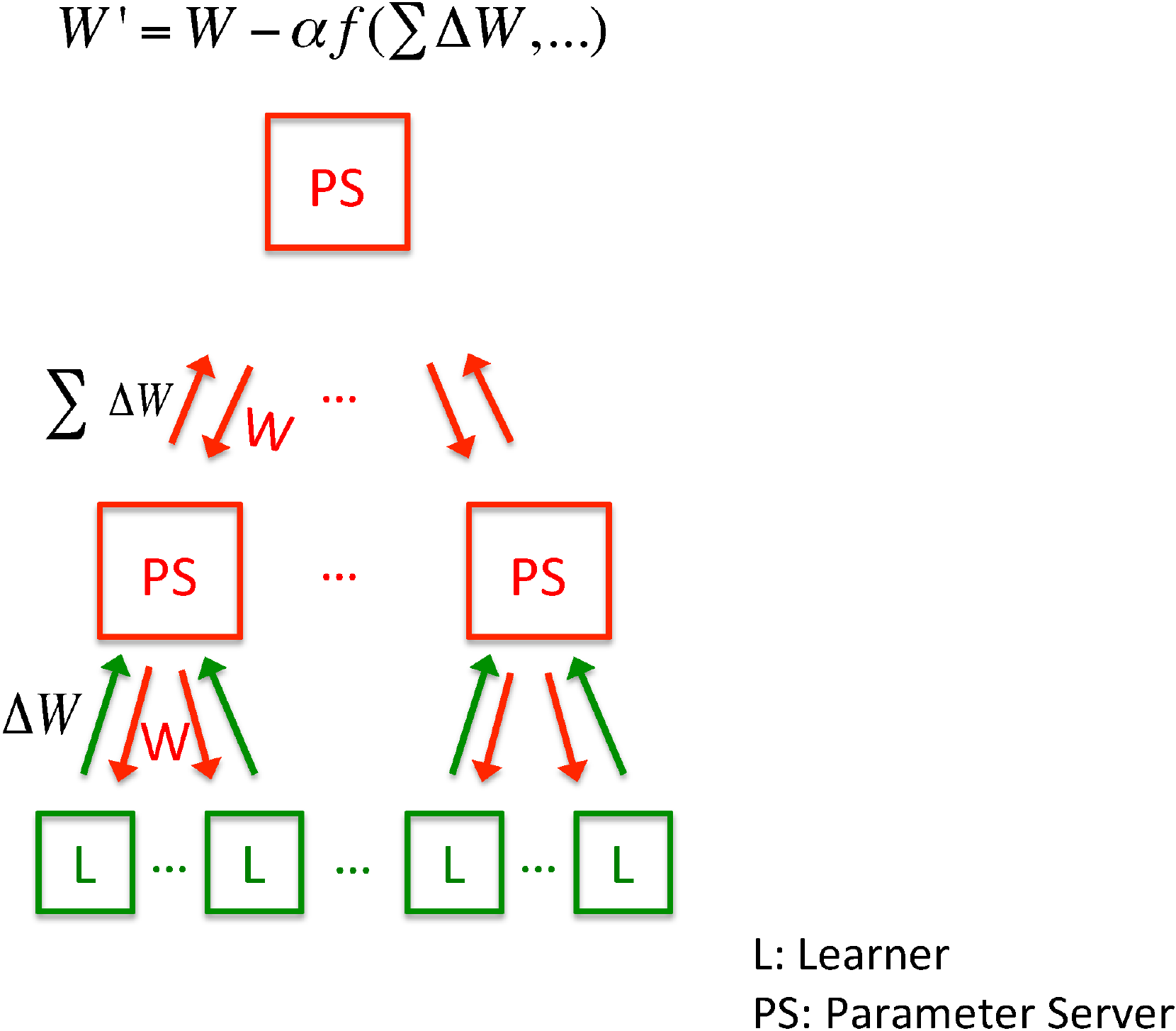}\label{fig:advstar_arch}}
\subfigure[\Advstar architecture]{\includegraphics[width=0.5\columnwidth]{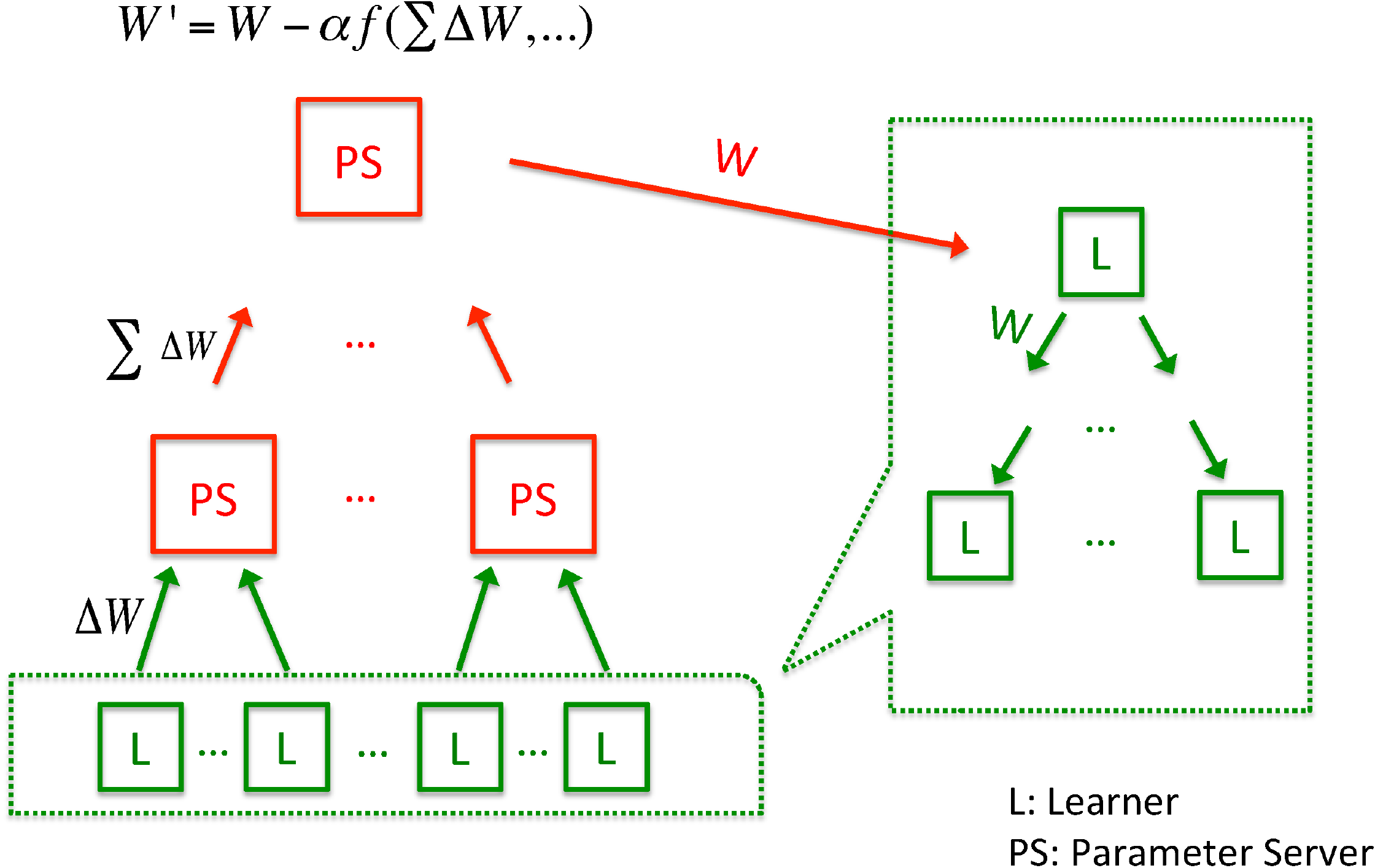}\label{fig:adv_arch} }
\vspace{-1mm}
\caption{\Adv architecture }
\vskip -0.1in
\end{figure}

\noindent\textbf{\Adv system architecture}. \Base clearly is not a scalable solution when the model gets large. Under ideal circumstances (see Section~\ref{sec:meth1} for our experimental hardware system specification), a single learner pushing a model of \SI{300}{\mega\byte} (size of a typical deep neural network, see section~\ref{sec:meth2}) would take more than \SI{10}{\milli\second} to transfer this data.  If 16 tasks are sending \SI{300}{\mega\byte} to the same receiver and there is link contention, it would take over a second for the messages to be delivered. 

To alleviate the network traffic to parameter server, we build a parameter server group that forms a tree.
We co-locate each tree leaf node on the same node as the learners for which it is responsible.
Each node in the parameter server group is responsible for averaging the gradients sent from its learners and relaying the averaged gradient to its parent.
The root node in the parameter server group is responsible for applying weight update and broadcasting the updated weights.
Each non-leaf node pulls the weights from its parent and responds to its children's weight pulling requests.
\Adv can significantly improve performance compared to \Base and manage to scale out to large model and small $\mu$, while maintaining the control of the gradients' staleness. Figure~\ref{fig:advstar_arch} illustrates the system architecture for \Adv. Red boxes represent the parameter server group, in which the gradients are pushed and aggregated upwards. Green boxes represent learners, each learner pushes the gradient to its parameter server parent and receives weights from its parameter server parent.
The key difference between \Adv and a sharded parameter server design (e.g., Distbelief~\cite{distbelief} and Adam~\cite{adam}) is that the weights maintained in \Adv have the same timestamps whereas shared parameter servers maintain the weights with different timestamps.
Having consistent weights makes the analysis of hyperparameter/scale-out interplay much more tractable.


\noindent\textbf{\Advstar system architecture}. We built \Advstar to further improve the runtime performance in two ways:

\underline{\textit{Broadcast weights within learners}}. To further reduce the traffic to the parameter server group, we form a tree within all learners and broadcast the weights down this tree. In this way the network links to/from learners are also utilized.

\underline{\textit{Asynchronous \pushGradient and \pullWeights}}. Ideally, one would use MPI non-blocking send calls to asynchronously send gradients and weights. However, depending on the MPI implementation, it is difficult to guarantee if the non-blocking send calls make progress in the background~\cite{mpi}.
Therefore we open additional communication threads and use MPI blocking send calls in the threads.
Each learner process runs two additional communication threads: the \pullWeights thread and \pushGradient thread.
In this manner, computing can continue without waiting for the communication. Note that since we need to control $\mu$ (the smaller $\mu$ is, the better model converges, as we demonstrate in Section~\ref{sec:tradeoffcurves}), we must guarantee that the learner pushes each calculated gradient to the server. Alternatively, one could locally accrue gradients and send the sum, as in~\cite{distbelief}, however that will effectively increase $\mu$.
For this reason, the \pushGradient thread cannot start sending the current gradient before the previous one has been delivered.
As demonstrated in Table~\ref{tab:comm-overlap} that as long as we can optimize the use of network links, this constraint has no bearing on the runtime performance, even when $\mu$ is extremely small.
In contrast, \pullWeights thread has no such constraint -- we maintain a computation buffer and a communication buffer for \pullWeights thread, and the communication always happens in the background.
To use the newly received weights only requires a pointer swap.
Figure~\ref{fig:adv_arch} illustrates the system architecture for \Advstar. Different from \Adv, each learner continuously receives weights from the weights downcast tree, which consists of the top level parameter server node and all the learners.

We measure the communication overlap by calculating the ratio between computation time and the sum of computation and communication time. Table~\ref{tab:comm-overlap} records the the communication overlap for \Base, \Adv, and \Advstar in an adversarial scenario. \Advstar can almost completely overlap computation with communication.
\Advstar can scale out to very large model size and work with smallest possible size of mini-batch.
In Section~\ref{eval:imgnet}, we demonstrate \Advstar's effectiveness in improving runtime performance while achieving good model accuracy.

\begin{table}
\centering{\small
    \begin{tabular}{|c|c|}
    \hline
    Implementation & Communication overlap (\%)  \\ 
    \hline
    \Base      &    11.52          \\
    \hline
    \Adv & 56.75 \\
    \hline
     \Advstar & 99.56 \\
    \hline
    \end{tabular}}
   \caption{Communication overlap measured in \Base, \Adv, \Advstar for an adversarial scenario, where the mini-batch size is the smallest possible for  4-way multi-threaded learners, model size 300MB, and there are about 60 leareners.}
  \label{tab:comm-overlap}
  \vskip -0.1in
\end{table}

\section{Methodology}
\label{sec:meth}
\subsection{\label{sec:meth1} Hardware and software environment}

We deploy the Rudra distributed deep learning framework on a P775 supercomputer.
Each node of this system contains four eight-core \SI{3.84}{\giga \Hz} POWER7 processors, one optical connect controller chip and \SI{128}{\giga \byte} of memory.
A single node has a theoretical floating point peak performance of \SI{982}{\giga flop/s}, memory bandwidth of \SI{512}{\giga \byte/s} and bi-directional interconnect bandwidth of \SI{192}{\giga\byte/\second}.

The cluster operating system is Red Hat Enterprise Linux 6.4.
To compile and run Rudra we used the IBM xlC compiler version 12.1 with the \texttt{-O3 -q64 -qsmp} options, ESSL for BLAS subroutines, and IBM MPI (IBM Parallel Operating Environment 1.2).
\subsection{\label{sec:meth2} Benchmark datasets and neural network architectures}
To evaluate Rudra's scale-out performance we employ two commonly used image classification benchmark datasets: \CIFAR \cite{krizhevsky2009learning} and \imagenet ~\cite{ILSVRC15}. 
The \CIFAR dataset comprises of a total of 60,000 RGB images of size 32 $\times$ 32  pixels partitioned into the training set (50,000 images) and the test set (10,000 images). 
Each image belongs to one of the 10 classes, with 6000 images per class. 
For this dataset, we construct a deep convolutional neural network (CNN) with 3 convolutional layers each followed by a subsampling/pooling layer. 
The output of the 3\ts{rd} pooling layer connects, via a fully-connected layer, to a 10-way softmax output layer that generates a probability distribution over the 10 output classes. 
This neural network architecture closely mimics the \CIFAR model (cifar10\_full.prototxt) available as a part of the open-source Caffe deep learning package \cite{jia2014caffe}. 
The total number of trainable parameters in this network are $\sim90$~K, resulting in the model size of $\sim$\SI{350}{\kilo \byte} when using 32-bit floating point data representation. 
The neural network is trained using momentum-accelerated mini-batch SGD with a batch size of 128 and momentum set to 0.9. 
As a data preprocessing step, the per-pixel mean is computed over the entire training dataset and subtracted from the input to the neural network. 
The training is performed for 140 epochs and results in a model that achieves 17.9\% misclassification error rate on the test dataset. 
The base learning rate $\alpha_0$ is set to 0.001 are reduced by a factor of 10 after the 120\ts{th} and 130\ts{th} epoch. 
This learning rate schedule proves to be quite essential in obtaining the low test error of 17.9\%.

Our second benchmark dataset is collection of natural images used as a part of the 2012 edition of the \imagenet Large Scale Visual Recognition Challenge (ILSVRC 2012). 
The training set is a subset of the hand-labeled \imagenet database and contains 1.2 million images. The validation dataset has 50,000 images. 
Each image maps to one of the 1000 non-overlapping object categories. The images are converted to a fixed resolution of 256$\times$256 to be used input to a deep convolution neural network.  
For this dataset, we consider the neural network architecture introduced in \cite{krizhevsky2012imagenet} consisting of 5 convolutional layers and 3 fully-connected layers. 
The last layer outputs the probability distribution over the 1000 object categories. In all, the neural network has $\sim$72 million trainable parameters and the total model size is \SI{289}{\mega \byte}. 
The network is trained using momentum-accelerated SGD with a batch size of 256 and momentum set to 0.9. Similar to the \CIFAR benchmark, per-pixel mean computed over the entire training dataset is subtracted from the input image feeding into the neural network. 
To prevent overfitting, a weight regularization penalty of 0.0005 is applied to all the layers in the network and a dropout of 50\% is applied to the 1\ts{st} and 2\ts{nd} fully-connected layers. 
The initial learning rate $\alpha_0$ is set equal to 0.01 and reduced by a factor of 10 after the 15\ts{th} and 25\ts{th} epoch. 
Training for 30 epochs results in a top-1 error of 43.95\% and top-5\footnote{The top-5 error corresponds to the fraction of samples where the correct label does not appear in the top-5 labels considered most probable by the model} error of 20.55\% on the validation set.
\begin{figure}[ht!]
\centering
\subfigure[\label{fig:staleness_a}]{
	\includegraphics[width=0.46\columnwidth]
	{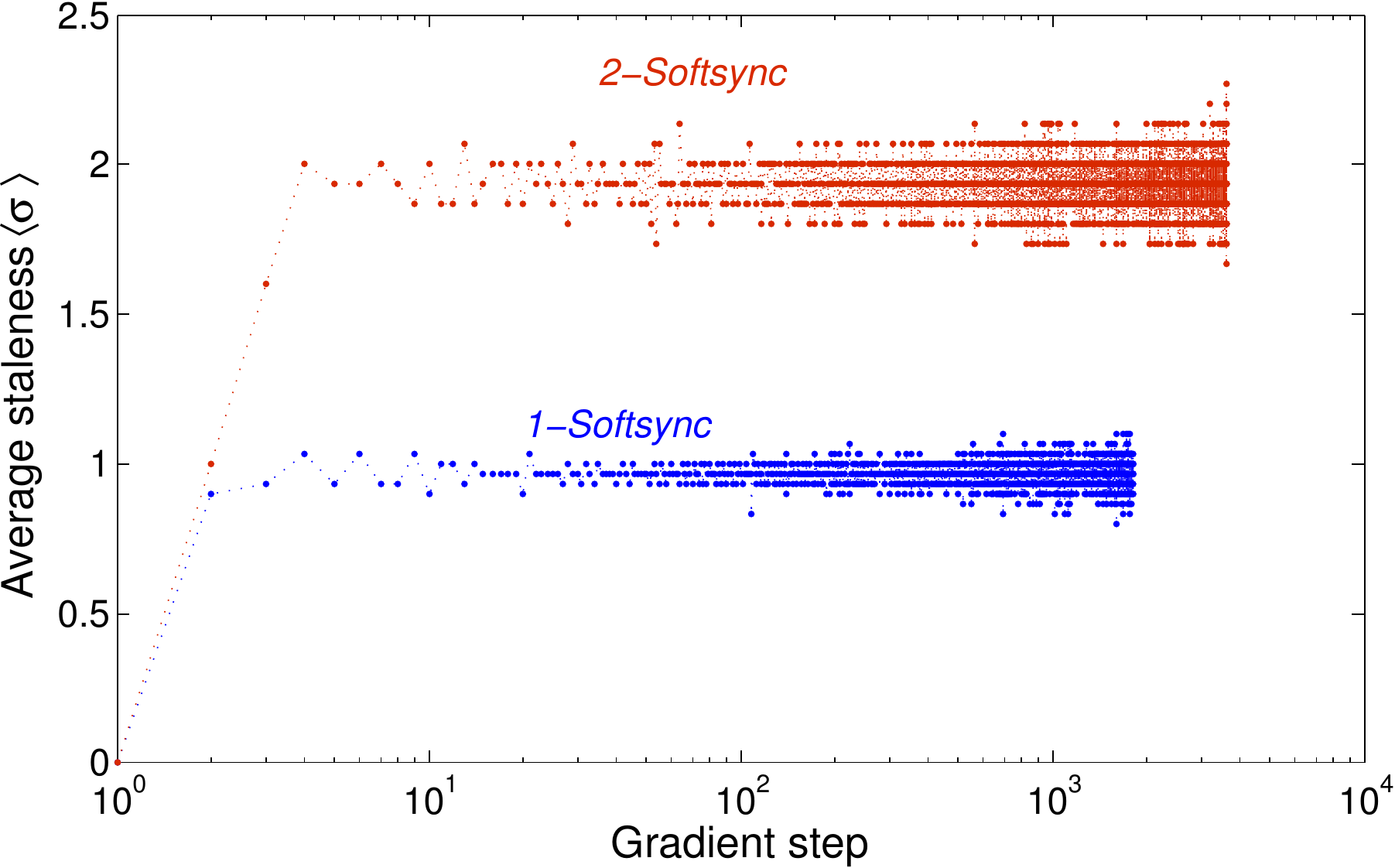}
}
\subfigure[\label{fig:staleness_b}]{
    \includegraphics[width=0.46\columnwidth]    
    {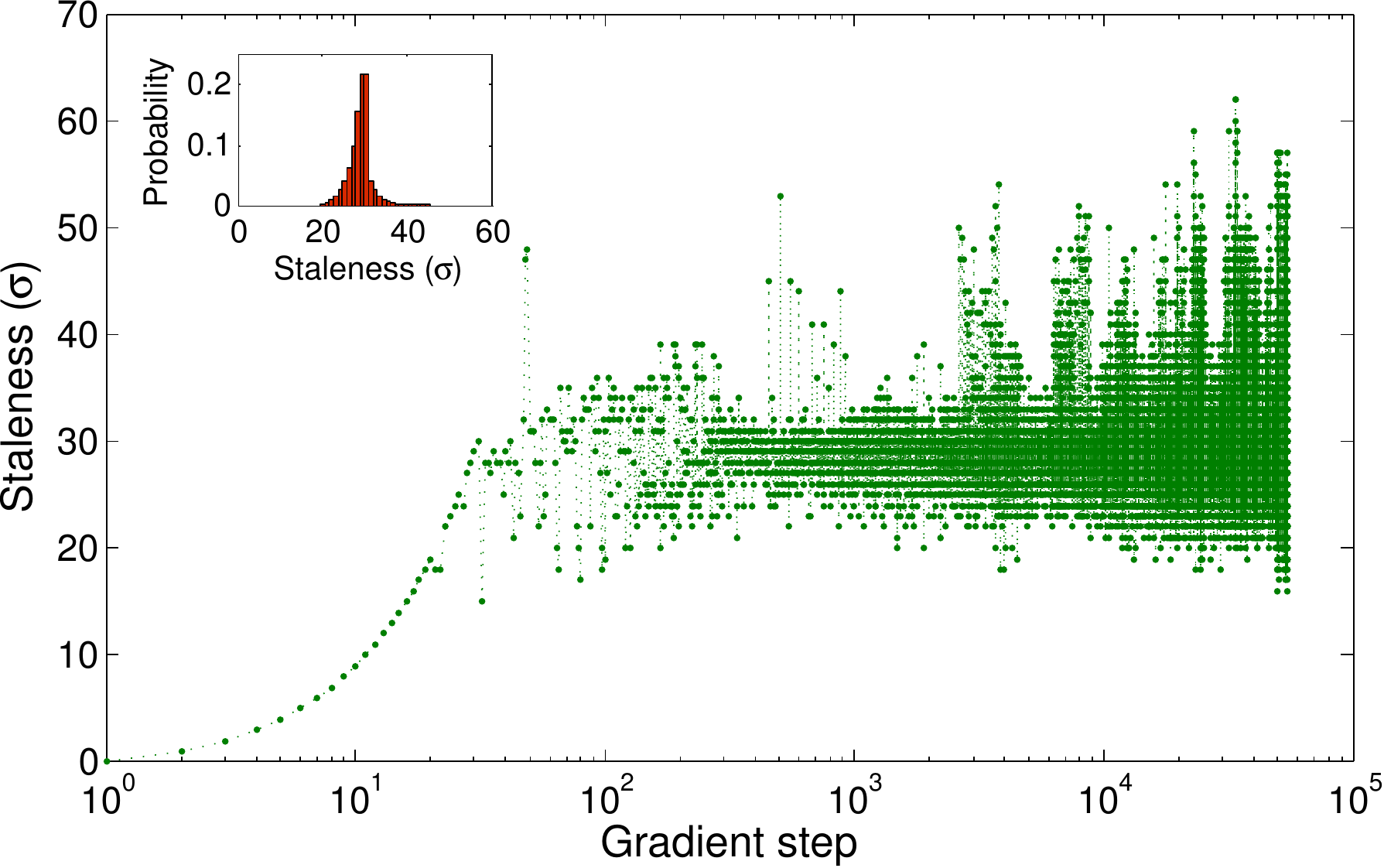}
}
\vspace{-1mm}
\caption{\label{fig:staleness} Average staleness $\langle\sigma\rangle$ of the gradients as a function of the weight update step at the parameter server when using (a)~1-softsync, 2-softsync and (b)~$\lambda$-softsync protocol. 
Inset in (b) shows the distribution of the gradient staleness values for $\lambda$-softsync protocol. Number of learners $\lambda$ is set to 30.}
\vskip -0.1in
\end{figure}

\section{Evaluation}
In this section we present results of evaluation of our scale-out deep learning training implementation. 
For an initial design space exploration, we use the \CIFAR dataset and \Base system architecture. 
Subsequently we extend our findings to the \imagenet dataset using the \Adv and \Advstar system architectures.
\subsection{\label{sec:stale_gradients}Stale gradients}
\label{sec:eval:staleness}

In the hardsync protocol introduced in section~\ref{sec:term_definitions}, the transition from $\theta(i)$ to $\theta(i+1)$ involves aggregating the gradients calculated with $\theta(i)$. 
As a result, each of the gradients $\nabla\theta_l$ carries with it a staleness $\sigma$ equal to 0. 
However, departing from the hardsync protocol towards either the $n$-softsync or the Async protocol inevitably adds staleness to the gradients, as a subset of the learners contribute gradients calculated using weights with timestamp earlier than the current timestamp of the weights at the parameter server.

To measure the effect of gradient staleness when using the $n$-softsync protocol, we use the \CIFAR dataset and train the neural network described in section~\ref{sec:meth2} using $\lambda = 30$ learners. 
For the $1$-softsync protocol, the parameter server updates the current set of weights when it has received a total of 30 gradients from the learners. 
Similarly, the $2$-softsync protocol forces the parameter server to accumulate $\lambda/2=15$ gradient contributions from the learners before updating the weights. 
As shown in Figure~\ref{fig:staleness_a} the average staleness $\langle\sigma\rangle$ for the 1-softsync and 2-softsync protocols remains close to 1 and 2, respectively. 
In the 1-softsync protocol, the staleness $\sigma_{L_{l}}$ for the gradients computed by the learner $L_l$ takes values 0, 1, or 2, whereas $\sigma_{L_{l}} \in \{0,1,2,3,4\}$ for the 2-softsync protocol. 
Figure~\ref{fig:staleness_b} shows the gradient staleness for the $n$-softsync protocol where $n=\lambda=30$. 
In this case, the parameter server updates the weights after receiving a gradient from any of the learners. 
A large fraction of the gradients have staleness close to 30, and only with a very low probability ($<0.0001$) does $\sigma$ exceed $2n=60$.
These measurements show that, in general,  $\sigma_{L_{l}} \in \{0,1,\hdots,2n\}$ and $\langle\sigma\rangle = n$ for our implementation of the $n$-softsync protocol.

\begin{figure}
\centering
\includegraphics[width=0.55\columnwidth]
{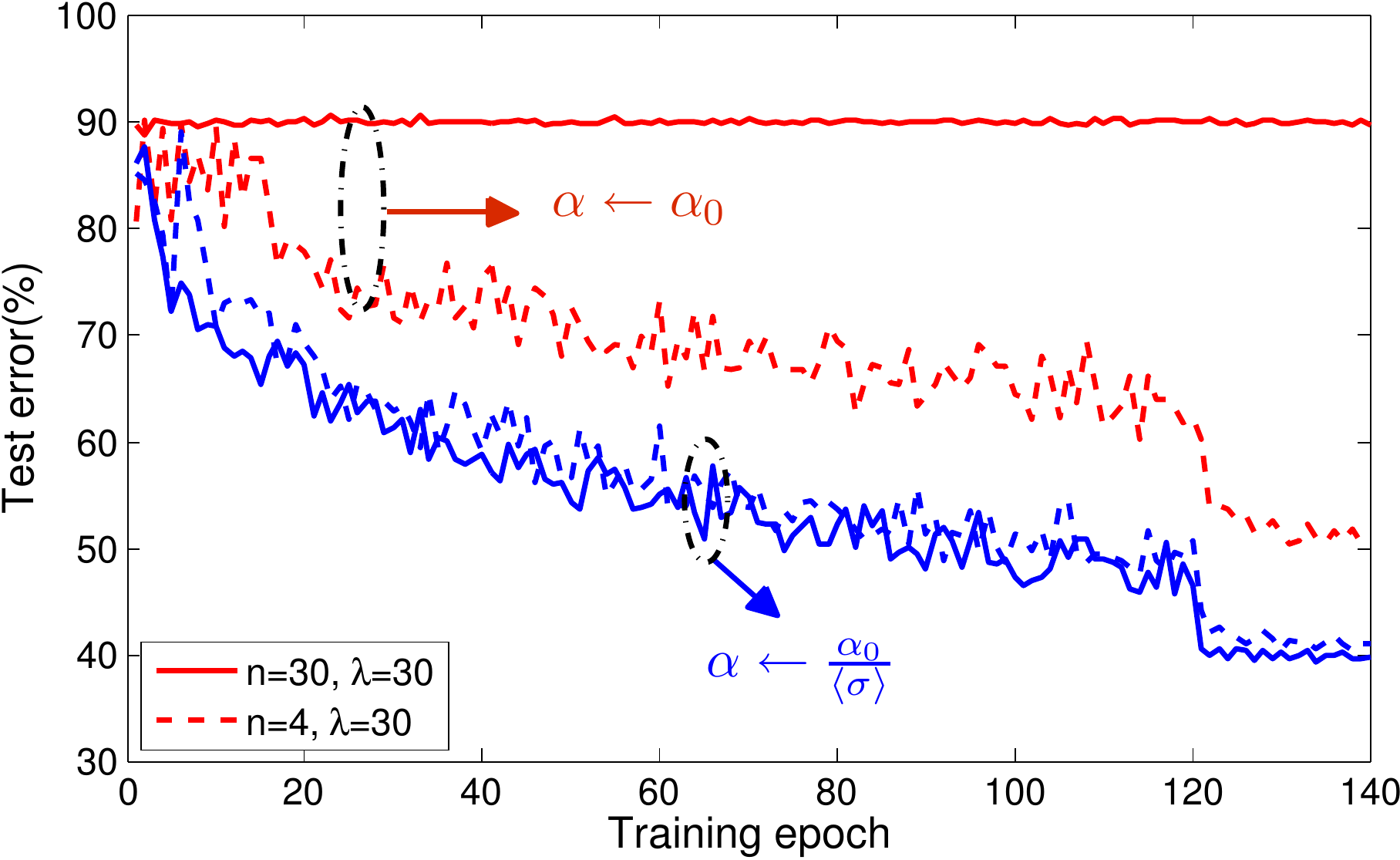}
\vspace{-2mm}
\caption{\label{fig:learningRate} \textit{Effect of learning rate modulation strategy}: Dividing the learning rate by the average staleness aids in better convergence and achieves lower test error when using the $n$-softsync protocol. Number of learners, $\lambda = 30$;  \minibatch size $\mu = 128$.}
\vskip -0.1in
\end{figure}

\underline{\textit{Modifying the learning rate for stale gradients}:} In our experiments with the $n$-softsync protocol we found it beneficial, and at times necessary, to modulate the learning rate $\alpha$ to take into account the staleness of the gradients. For the $n$-softsync protocol, we set the learning rate as: {\small
\begin{equation}
\alpha = {\alpha_0}/{\langle\sigma\rangle} = {\alpha_0}/{n} 
\label{eq:learningrate_eq}
\end{equation}}where $\alpha_0$ is the learning rate used for the baseline (control) experiment: $\mu=128$, $\lambda=1$.
Figure~\ref{fig:learningRate} shows a set of representative results illustrating the benefits of adopting this learning rate modulation strategy. 
We show the evolution of the test error on the \CIFAR dataset as a function of the training epoch for two different configurations of the $n$-softsync protocol ($n = 4$, $n = 30$) and set the number of learners, $\lambda=30$. 
In both these configurations, setting the learning rate in accordance with equation~\eqref{eq:learningrate_eq} results in lower test error as compared with the cases where the learning rate is set to $\alpha_0$. 
Surprisingly, the configuration $30$-softsync, $\lambda=30$, $\alpha=\alpha_0$ fails to converge and shows a constant high error rate of 90\%. 
Reducing the learning rate by a factor $\langle\sigma\rangle = n = 30$ makes the model with much lower test error\footnote{Although not explored as a part of this work, it is certainly possible to implement a finer-grained learning rate modulation strategy that depends on the staleness of each of gradients computed by the learners instead of the average staleness. Such a strategy should apply smaller learning rates to staler gradients}.

\subsection{$(\sigma,\mu,\lambda)$ tradeoff curves}
\label{sec:tradeoffcurves}
Hyperparameter optimization plays a central role in obtaining good accuracy from neural network models \cite{breuel2015effects}.
For the SGD training algorithm, this includes a search over the neural network's training parameters such as learning rates, weight regularization, depth of the network, \minibatch size etc.\xspace in order to improve the quality of the trained neural network model (quantified as the error on the validation dataset). 
Additionally, when distributing the training problem across multiple learners, parameters such as the number of learners and the synchronization protocol enforced amongst the learners impact not only the runtime of the algorithm but also the quality of the trained model.

\begin{figure}
 \vskip -0.0in
 \centering
    	\includegraphics[width=0.55\columnwidth,height=0.35\columnwidth]    {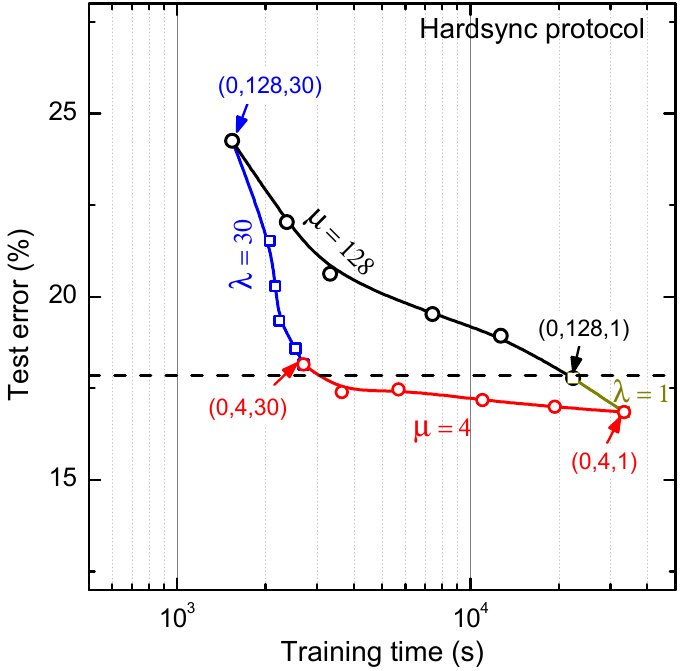} 
    	\vspace{-2mm}
    \caption{\label{fig:SML_hard}\SML tradeoff curves for the hardsync protocol. The dashed black line represents the 17.9\% test error achieved by the baseline model \SML $=(0,128,1)$ on the \CIFAR dataset.}
\vskip -0.1in
\end{figure}

An exhaustive search over the space defined by these parameters for joint optimization of the runtime performance and the model quality can prove to be a daunting task even for a small model such as that used for the \CIFAR dataset, and clearly intractable for models and datasets the scale of \imagenet. 
To develop a better understanding of the interdependence among the various tunable parameters in the distributed deep learning problem, we introduce the notion of \SML tradeoff curves. 
A \SML tradeoff curve plots the error on the validation set (or test set) and the total time to train the model (wall clock time) for different configurations of average gradient staleness $\langle\sigma\rangle$, \minibatch size per learner $\mu$, and the number of learners $\lambda$. 
The configurations \SML that achieve the virtuous combination of low test error and small training time are preferred and form ideal candidates for further hyperparameter optimization.

We run\footnote{The mapping between $\lambda$ and the number of computing nodes $\eta$ is $(\lambda,\eta)=\{(1,1),(2,1),(4,2),(10,4),(18,4),(30,8)\}$} the \CIFAR benchmark for $\lambda~\in~\{1,2, 4,10,18,30\}$ and $\mu \in \{4,8,16,32,64,128\}$. Figure~\ref{fig:SML_hard} shows a set of \SML curves for the hardsync protocol i.e.\ $\sigma = 0$. 
The baseline configuration with $\lambda=1$ learner and \minibatch size $\mu = 128$ achieves a test error of 17.9\%. 
With the exception of modifying the learning rate as $\alpha = \alpha_0\sqrt{\mu \lambda / 128}$, all the other neural network's hyperparameters were kept unchanged from the baseline configuration while generating the data points for different values of $\mu$ and $\lambda$. 
Note that it is possible to achieve test error lower than the baseline by reducing the \minibatch size from 128 to 4. 
However, this configuration (indicated on the plot as \SML~$=(0,4,1)$) increases training time compared with the baseline. 
This is primarily due to the fact that the dominant computation performed by the learners involves multiple calls to matrix multiplication (GEMM) to compute $WX$ where samples in a \minibatch form columns of the matrix $X$. 
Reducing the \minibatch size cause a proportionate decrease in the GEMM throughput and slower processing of the \minibatch by the learner. 
\begin{figure}
 \centering
	\subfigure[\label{fig:SML_async}]{
	 \includegraphics[width=0.46\columnwidth]
	 {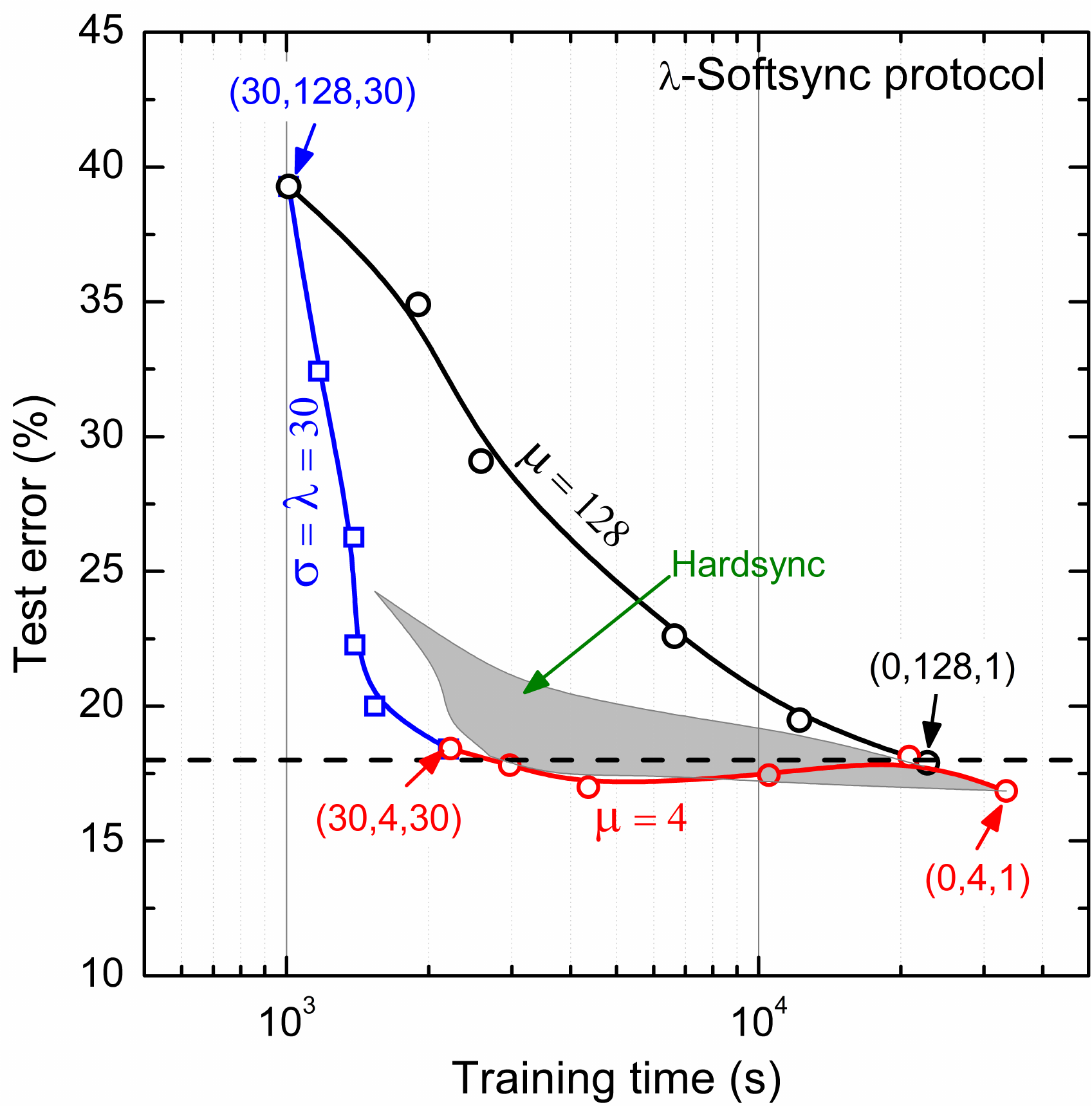}
}
	\subfigure[\label{fig:SML_smart}]{
	 \includegraphics[width=0.46\columnwidth]
	 {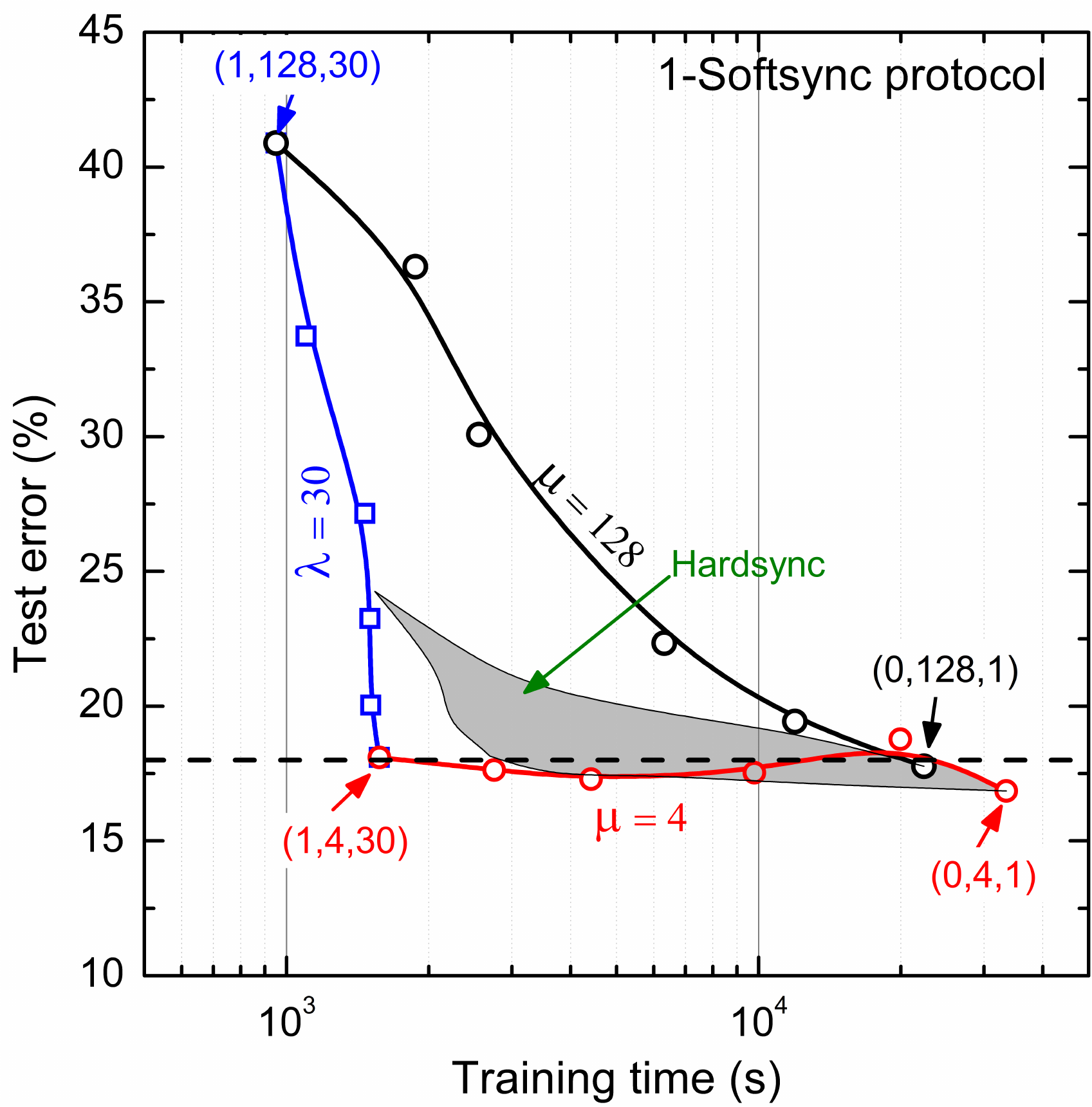}
}
\vspace{-2mm}
\caption{\label{fig:SMLcurves} \SML tradeoff curves for (a) $\lambda$-softsync protocol and (b) $1$-softsync protocol. Shaded region in  shows the region bounded by $\mu=128$, $\lambda=30$, and $\mu=4$ contours for the hardsync protocol. $\lambda~\in~\{1,2, 4,10,18,30\}$ and $\mu \in \{4,8,16,32,64,128\}$. Note that for $\lambda=1$, $n$-softsync protocol degenerates to the hardsync protocol}
\end{figure}

In Figure~\ref{fig:SML_hard}, the contour labeled  $\mu=128$ is the configurations with the \minibatch size per learner is kept constant at 128 and the number of learners is varied from $\lambda = 1$ to $\lambda=30$. 
The training time reduces monotonically with $\lambda$, albeit at the expense of an increase in the test error. 
Traversing along the $\lambda=30$ contour from configuration \SML~$=(0,128,30)$ to \SML~$=(0,4,30)$ (i.e.\ reducing the \minibatch size from 128 to 4) helps restore much of this degradation in the test error by partially sacrificing the speed-up obtained by the virtue of scaling out to 30 learners. 

\begin{figure}[ht]
\vspace{2mm}
\centering
\subfigure[\small]{
	\includegraphics[width=0.46\columnwidth]
{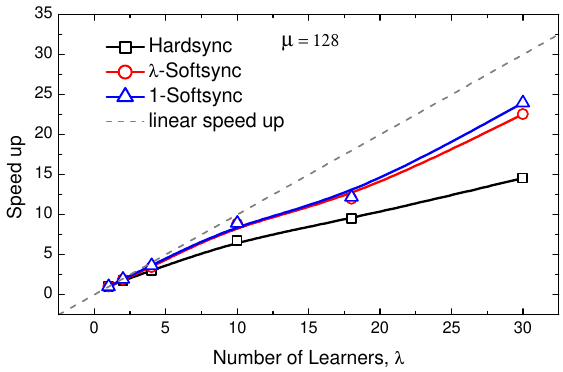}
}
\subfigure[\small]{
    \includegraphics[width=0.46\columnwidth]    {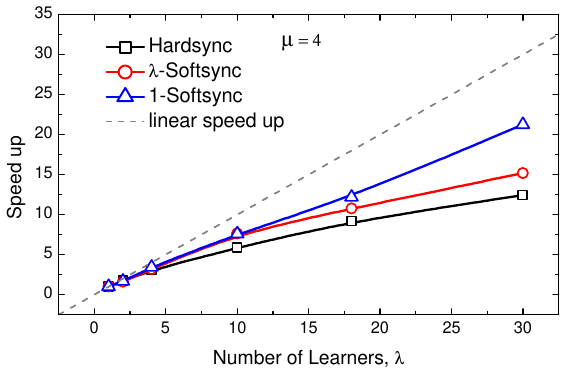}
}
\vspace{-2mm}
\caption{\label{fig:speedup} Speed-up in the training time for \minibatch size and (a) $\mu=128$ (b) $\mu=4$  for 3 different protocols: hardsync, $\lambda$-softsync, and $1$-softsync. Speed-up numbers in (a) and (b) are calculated relative to \SML~$=(0,128,1)$ and \SML~$=(0,4,1)$, respectively.}
\end{figure}
Figure~\ref{fig:SML_async} shows \SML tradeoff curves for the $\lambda$-softsync protocol. 
In this protocol, the parameter server updates the weights as soon as it receives a gradient from any of the learners. 
Therefore, as shown in section~\ref{sec:stale_gradients} the average gradient staleness $\langle\sigma\rangle = \lambda$ and $\sigma_{max} \leq 2\lambda$ with high probability. 
The learning rate is set in accordance with equation~\ref{eq:learningrate_eq}. 
All the other hyperparameters are left unchanged from the baseline configuration. 
Qualitatively, the \SML tradeoff curves for $\lambda$-softsync look similar to those observed for the hardsync protocol. 
In this case, however, the degradation in the test error relative to the baseline for the \SML~$=(30,128,30)$ configuration is much more pronounced. 
As observed previously, this increase in the test error can largely be mitigated by reducing the size of \minibatch processed by each learner ($\lambda=30$ contour). 
Note, however, the sharp increase in the training time for the configuration \SML~$=(30,4,30)$ as compared with \SML~$=(30,128,30)$. 
The smaller \minibatch size not only reduces the computational throughput at each learner, but also increases the frequency of \pushGradient and \pullWeights requests at the parameter server. 
In addition, small \minibatch size increases the frequency of weight updates at the parameter server. 
Since in the \Base architecture (section~\ref{sec:base_arch}), the learner does not proceed with the computation on the next \minibatch till it has received the updated gradients, the traffic at the parameter server and the more frequent weight updates can delay servicing the learner's \pullWeights request, potentially stalling the gradient computation at the learner. 
Interestingly, all the configurations along the $\mu=4$ contour show similar, if not better, test error as the baseline. For these configurations, the average staleness varies between 2 and 30. \textit{From this empirical observation, we infer that a small \minibatch size per learner confers upon the training algorithm a fairly high degree of immunity to stale gradients.} 

The $1$-softsync protocol shows \SML tradeoff curves (Figure~\ref{fig:SML_smart}) that appear nearly identical to those observed for the $\lambda$-softsync protocol. 
In this case, the average staleness is 1 irrespective of the number of learners. 
Since the parameter server waits for $\lambda$ gradients to arrive before updating the weights, there is a net reduction in the \pullWeights traffic at the parameter server (see section~\ref{sec:base_arch}). 
As a result, the $1$-softsync protocol avoids the degradation in runtime observed in the $\lambda$-softsync protocol for the configuration with $\mu=4$ and $\lambda=30$. 
The distinction in terms of the runtime performance between these two protocols becomes obvious when comparing the speed-ups obtained for different \minibatch sizes (Figure~\ref{fig:speedup}). 
For $\mu=128$, the $1$-softsync and $\lambda$-softsync protocol demonstrate similar speed-ups over the baseline configuration for upto $\lambda=30$ learners. 
In this case, the communication between the learners and the parameter server is sporadic enough to prevent the learners from waiting on the parameter server for updated weights. 
However, as the number of learners is increased beyond 30, the bottlenecks at the parameter server are expected to diminish the speed-up obtainable using the $\lambda$-softsync protocol. 
The effect of frequent \pushGradient and \pullWeights requests due to smaller at the parameter manifest clearly as the \minibatch size is reduced to 4, in which case, the $\lambda$-softsync protocol shows subdued speed-up compared with $1$-softsync protocol. 
In either scenario, the hardsync protocol fares the worst in terms of runtime performance improvement when scaling out to large number of learners. 
The upside of adopting the hardsync protocol, however, is that it achieves substantially lower test error, even for large \minibatch sizes.

\subsection{\label{sec:mulambda}$\mu\lambda = \text{constant}$}

In the hardsync protocol, given a configuration with $\lambda$ learners and \minibatch size $\mu$ per learner, the parameter server averages the $\lambda$ number of gradients reported to it by the learners. Using equations~\eqref{eq:gd} and \eqref{eqn:update_hard}:{\small
\begin{eqnarray}
		\nabla \theta^{(k)}(i) &=& \frac{1}{\lambda}{\sum_{l=1}^{\lambda} \nabla\theta_{l}^{(k)}}=\frac{1}{\lambda}\sum_{l=1}^{\lambda}\left(\frac{1}{\mu}\sum_{s=1}^{\mu}\frac{\partial C \left(\hat{Y_s},Y_s\right)}{\partial \theta_{l}^{(k)}(i)}\right) \nonumber\\
		&=&\frac{1}{\mu\lambda}\sum_{s=1}^{\mu\lambda}\frac{\partial C \left(\hat{Y_s},Y_s\right)}{\partial \theta^{(k)}(i)} \label{eqn:mulambda_const}
\end{eqnarray}}The last step equation~\eqref{eqn:mulambda_const} is valid since each training example $(X_s,Y_s)$ is drawn independently from the training set and also because the hardsync protocol ensures that all the learners compute gradients on identical set of weights i.e. $ \theta_{l}^{(k)}(i) = \theta^{(k)}(i)$~$\forall$~ $l \in \{1,2,\hdots,\lambda\}$. 
According to equation~\eqref{eqn:mulambda_const}, the configurations \SML~$=(0,\mu_0\lambda_0,1)$ and \SML~$=(0,\mu_0,\lambda_0)$ are equivalent from the perspective of stochastic gradient descent optimization. 
In general, hardsync configurations with the same $\mu\lambda$ product are expected to give nearly\footnote{small differences in the final test error achieved may arise due to the inherent nondeterminism of random sampling in stochastic gradient descent and the random initialization of the weights.} the same test error.
\newcolumntype{g}{>{\columncolor{Gray}}c}
\begin{table}[!ht]
  \centering{tiny
   \begin{tabular}{c|cccgc}
    \hline
    & $\sigma$ & $\mu$ & $\lambda$ & \multicolumn{1}{l}{\begin{tabular}[c]{@{}c@{}}Test \\ error \end{tabular}}  & \begin{tabular}[c]{@{}c@{}}Training \\ time(s)\end{tabular} \\ 
    \hline    
    & \textbf{1}       & \textbf{4}   &  \textbf{30} & \textbf{18.09}\%         & \textbf{1573} \T   \\
    & 30      &  4  &  30 & 18.41\%         & 2073  \\         
    $\mu\lambda \approx 128 $&18 & 8 & 18 & 18.92\%        & 2488 \\
    &10 	    & 16  &  10 & 18.79\%          & 3396  \\
    & 4      &  32  &  4 & 18.82\%         & 7776  \\  
    & 2     &  64  &  2 & 17.96\%         & 13449  \B \\  
    \hline
    & \textbf{1}  & \textbf{8} & \textbf{30} & \textbf{20.04}\% & \textbf{1478} \T \\
    & 30 & 8 & 30 & 19.65\% & 1509 \\ 
    & 18 & 16 & 18 & 20.33\% & 2938 \\
    $\mu\lambda \approx 256 $		  & 10 & 32 & 10 & 20.82\% & 3518 \\
    & 4 & 64 & 4 & 20.70\% & 6631   \\
    & 2 & 128 & 2 & 19.52\% & 11797 \\
    & 1  & 128 & 2 & 19.59\% & 11924 \B \\
    \hline
    & \textbf{1}  & \textbf{16} & \textbf{30} & \textbf{23.25}\% & \textbf{1469} \T \\
    & 30 & 16 & 30 & 22.14\% & 1502 \\
    $\mu\lambda \approx 512 $ & 18 & 32 & 18 & 23.63\% & 2255 \\
    & 10 & 64 & 10 & 24.08\% & 2683 \\
    & 4  & 128 & 4 & 23.01\% & 7089 \B \\
    \hline 
    & \textbf{1}  & \textbf{32} & \textbf{30} & \textbf{27.16}\% & \textbf{1299} \T \\
    & 30 & 32 & 30 & 27.27\% & 1420 \\
    $\mu\lambda \approx 1024 $ & 18 & 64 & 18 & 28.31\% & 1713 \\
    & 1 & 128 & 10 & 29.83\% & 2551 \\
    & 10 & 128 & 10 & 29.90\% & 2626 \\
    \hline 		
   \end{tabular}}
   \caption{\textit{Results on \CIFAR benchmark}: Test error at the end of 140 epochs and training time for \SML configurations with $\mu\lambda=\text{constant}$. }
   
    \label{tab:mullambda}
 \end{table} 

In Table~\ref{tab:mullambda} we report the test error at the end of 140 epochs for configurations with $\mu\lambda = \text{constant}$.
Interestingly, we find that even for the $n$-softsync protocol, configurations that maintain $\mu\lambda = \text{constant}$ achieve comparable test errors.
At the same time, the test error turns out to be rather independent of the staleness in the gradients for a given $\mu\lambda$ product. 
For instance, Table~\ref{tab:mullambda} shows that when $\mu\lambda \approx 128$, but the (average) gradient staleness is allowed to vary between 1 and 30, the test error stays $\sim$18-19\%.
Although this result may seem counter-intuitive, a plausible explanation emerges when considering the measurements shown earlier in Figure~\ref{fig:staleness}, that our implementation of the $n$-softsync protocol achieves an average gradient staleness of $n$ while bounding the maximum staleness at $2n$.
Consequently, at any stage in the gradient descent algorithm, the weights being used by the different learners ($\theta_{l}(t)$) do not differ significantly and can be considered to be approximately the same. The quality of this approximation improves when each update \[\theta^{(k)}(i+1) = \theta^{(k)}(i) - \alpha\nabla \theta^{(k)}(i)\] creates only a small displacement in the weight space. 
This can be controlled by suitably tuning the learning rate $\alpha$. Qualitatively, the learning rate should decrease as the staleness in the system increases in order to reduce the divergence across the weights seen by the learners. The learning rate modulation of equation~\eqref{eq:learningrate_eq} achieves precisely this effect.

These results help define a principled approach for distributed training of neural networks: \emph{the \minibatch size per learner should be reduced as more learners are added to the system in way that keeps $\mu\lambda$ product constant. In addition, the learning rate should be modulated to account for stale gradients.}
In Table~\ref{tab:mullambda}, $1$-softsync ($\sigma = 1$) protocol invariably shows the smallest training time for any $\mu\lambda$. This is expected, since the $1$-softsync protocol minimizes the traffic at the parameter server. 
Table~\ref{tab:mullambda} also shows that the test error increases monotonically with the $\mu\lambda$ product.  These results reveal the scalability limits under the constraints of preserving the model accuracy. 
Since the smallest possible \minibatch size is 1, the maximum number of learners is bounded. 
This upper bound on the maximum number of learners can be relaxed if an inferior model is acceptable. Alternatively, it may be possible to reduce the test error for higher $\mu\lambda$ by running for more number of epochs. 
In such a scenario, adding more learners to the system may give diminishing improvements in the overall runtime. From machine learning perspective, this points to an interesting research direction on designing optimization algorithm and learning strategies that perform well with large \minibatch sizes.

\subsection{Summary of results on \CIFAR benchmark}
Table~\ref{tab:top-5} summarizes the results obtained on the \CIFAR dataset using the \Base system architecture. As a reference for comparison, the baseline configuration \SML~$=(0,128,1)$ achieves a test error of 17.9\% and takes 22,392 seconds to finish training 140 epochs.

\begin{table}
\centering
{\small
    \begin{tabular}{cccccc}
    \hline
    $\sigma$ & $\mu$ & $\lambda$ & \multicolumn{1}{l}{\begin{tabular}[c]{@{}c@{}}Synchronization \\ protocol \end{tabular}} & \begin{tabular}[c]{@{}c@{}}Test \\ error\end{tabular}& \begin{tabular}[c]{@{}c@{}}Training \\ time(s)\end{tabular} \\ 
    \hline
    1      & 4      &     30 & $1$-softsync			&\textbf{18.09}\%         & \textbf{1573}          \T   \\ 
    0      & 8      &     30 & Hardsync 				& 18.56\%         & 1995             \\ 
    30     & 4      &     30 & $30$-softsync			& 18.41\%         & 2073             \\ 
    0      & 4      &     30 & Hardsync				& 18.15\%         & 2235             \\ 
    18     & 8      &     18 & $18$-softsyc			& 18.92\%         & 2488            \B \\ 
    \hline
    \end{tabular}}
   \caption{\textit{Results on \CIFAR benchmark}: Top-5 best \SML configurations that achieve a combination of low test error and small training time.}
  \label{tab:top-5}
\end{table}

\begin{table*}
\centering{\small
    \begin{tabular}{llcccccc}
    \hline
  Configuration &Architecture & $\mu$ & $\lambda$ &\multicolumn{1}{l}{\begin{tabular}[c]{@{}c@{}}Synchronization \\ protocol \end{tabular}} & \begin{tabular}[c]{@{}c@{}}Validation \\ error(top-1) \end{tabular}  & \begin{tabular}[c]{@{}c@{}}Validation \\ error (top-5) \end{tabular} & \begin{tabular}[c]{@{}c@{}}Training time\\ (minutes/epoch) \end{tabular} \\ 
    \hline
	\cfga & \Base & 16 & 18  & Hardsync & 44.35\% & 20.85\% & 330 \T \\
	\cfgb & \Base & 16 & 18 & $1$-softsync & 45.63\% & 22.08\% & 270 \\ 
	\cfgc &	\Adv & 4 & 54 & $1$-softsync & 46.09\% &  22.44\% & 212 \\
	\cfgd & \Advstar  & 4 & 54 & $1$-softsync & 46.53\% & 23.38\% & 125 \B \\
    \hline
    \end{tabular}}
   \caption{\textit{Results on \imagenet benchmark}: Validation error at the end of 30 epochs and training time per epoch for different configurations.}
  \label{tab:imagenet}
\end{table*}

\subsection{Results on \imagenet benchmark}
\label{eval:imgnet}
The large model size of the neural network used for the \imagenet benchmark and the associated computational cost of training this model prohibits an exhaustive state space exploration. The baseline configuration ($\mu=256$, $\lambda=1$) takes 54 hours/epoch.
Guided by the results of section~\ref{sec:mulambda}, we first consider a configuration with $\mu=16$, $\lambda=18$ and employ the \Base architecture with hardsync protocol (\cfga). 
This configuration performs training at the speed of $\sim$330 minutes/epoch and achieves a top-5 error of 20.85\%, matching the accuracy of the baseline configuration ($\mu=256$, $\lambda=1$, section \ref{sec:meth2}). 

The synchronization overheads associated with the hardsync protocol deteriorate the runtime performance and the training speed can be further improved by switching over to the $1$-softsync protocol. 
Training using the $1$-softsync protocol with \minibatch size of $\mu=16$ and 18 learners takes $\sim$270 minutes/epoch, reaching a top-1 (top-5) accuracy of 45.63\% (22.08\%) by the end of 30 epochs (\cfgb).
For this particular benchmark, the training setup for the $1$-softsync protocol differs from the hardsync protocol in certain subtle, but important ways. 
First, we use an adaptive learning rate method (AdaGrad \cite{duchi2011adaptive,distbelief}) to improve the stability of SGD when training using the $1$-softsync protocol. 
Second, to improve convergence we adopt the strategy of warmstarting \cite{senior2013empirical} the training procedure by initializing the network's weights from a model trained with hardsync for 1 epoch. 

Further improvement in the runtime performance may be obtained by adding more learners to the system. 
However, as observed in the previous section, increase in the number of learners needs to be accompanied by a reduction in the \minibatch size to prevent degradation in the accuracy of the trained model. 
The combination of a large number of learners and a small \minibatch size represents a scenario where the \Base architecture may suffer from a bottleneck at the parameter server due to the frequent \pushGradient and \pullWeights requests. 
These effects are expected to be more pronounced for large models such as \imagenet. 
The \Adv architecture alleviates these bottlenecks, to some extent, by implementing a parameter server group organized in a tree structure.  $\lambda=54$ learners, each processing a \minibatch size $\mu=4$ trains at $\sim$212 minutes/epoch when using \Adv architecture and $1$-softsync protocol (\cfgc).
As in the case of \Base, the average staleness in the gradients is close to 1 and this configuration also achieves a top-1(top-5) error of 46.09\%(22.44\%).

The \Advstar architecture improves the runtime further by preventing the computation at the learner from stalling on the parameter server. 
However, this improvement in performance comes at the cost of increasing the average staleness in the gradients, which may deteriorate the quality of the trained model. 
The previous configuration runs at $\sim$125 minutes/epoch, but suffers an increase in the top-1 validation error (46.53\%) when using \Advstar architecture (\cfgd). Table~\ref{tab:imagenet} summarizes the results obtained for the 4 configurations discussed above. 
It is worth mentioning that the configuration $\mu=8$, $\lambda=54$ performs significantly worse, producing a model that gives top-1 error of $>$ 50\% but trains at a speed of $\sim$96 minutes/epoch. 
This supports our observation that scaling out to large number of learners must be accompanied by reducing the \minibatch size per learner so the quality of the trained model can be preserved.  

\begin{figure}[tbp]
  \centering
  \includegraphics[width=0.65\columnwidth]
  {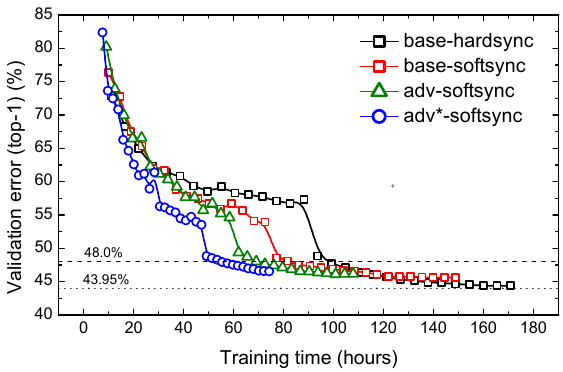}
  \vspace{-1mm}
  \caption{\label{fig:imagenetStory}\textit{Results on \imagenet benchmark:} Error on the validation set as a function of training time for the configurations listed in Table~\ref{tab:imagenet}}
\vskip -0.1in
\end{figure}

Figure~\ref{fig:imagenetStory} compares the evolution of the top-1 validation error during training for the 4 different configuration summarized in Table~\ref{tab:imagenet}. The training speed follows the order $\cfgd > \cfgc > \cfgb > \cfga$. As a result, \cfgd is the first configuration to hit the 48\% validation error mark. Configurations other than \cfga show marginally higher validation error compared with the baseline. As mentioned earlier, the experiments with $1$-softsync protocol use AdaGrad to achieve stable convergence. It is well-documented \cite{zeiler2012adadelta,senior2013empirical} that AdaGrad is sensitive to the initial setting on the learning rates. We speculate that tuning the initial learning rate can help recover the slight loss of accuracy for the $1$-softsync runs.

\section{Related Works}
\label{sec:scal-out-dl}

To accelerate training of deep neural networks and handle large dataset and model size, many researchers have adopted GPU-based solutions, either single-GPU~\cite{krizhevsky2012imagenet} or multi-GPU~\cite{mariana}
GPUs provide enormous computing power and are particularly suited for the matrix multiplications which are the core of many deep learning implementations.
However, the relatively limited memory available on GPUs may restrict their applicability to large model sizes.

Distbelief~\cite{distbelief} pioneered scale-out deep learning on CPUs. Distbelief is built on tens of thousands of commodity PCs and employs both model parallelism via dividing a single model between learners, and data parallelism via model replication.
To reduce traffic to the parameter server, Distbelief shards parameters over a parameter server group.
Learners asynchronously push gradients and pull weights from the parameter server.
The frequency of communication can be tuned via \textit{npush} and \textit{nfetch} parameters.

More recently, Adam~\cite{adam} employs a similar system architecture to DistBelief, while improving on Distbelief in two respects: (1) better system tuning, e.g. customized concurrent memory allocator, better linear algebra library implementation, and passing activation and error gradient vector instead of the weights update; and (2) leveraging the recent improvement in machine learning, in particular convolutional neural network to achieve better model accuracy.


In any parameter server based deep learning system~\cite{staleness}, staleness will negatively impact model accuracy. Orthogonal to the system design, many researchers have proposed solutions to counter staleness in the system, such as bounding the staleness in the system~\cite{bounded-staleness} or changing optimization objective function, such as elastic averaging SGD~\cite{elastic-averaging-sgd}. 
In this paper, we empirically study how staleness affects the model accuracy and discover the heuristics that smaller \minibatch size can effectively counter system staleness. In our experiments, we derive this heuristics from a small problem size(e.g., \CIFAR) and we find this heuristic is applicable even to much larger problem size (e.g., \imagenet). Our finding coincides with a very recent theoretical paper~\cite{liu-asgd-nips-2015}, in which the authors prove that in order to achieve linear speedup using asynchronous protocol while maintaining good model accuracy, one needs to increase the number of weight updates conducted at the parameter server. In our system, this theoretical finding is equivalent to keeping constant number of training epochs while reducing the \minibatch size.
To the best of our knowledge, our work is the first systematic study of the tradeoff between model accuracy and runtime performance for distributed deep learning.

\section{Conclusion}
In this paper, we empirically studied the interplay of hyper-parameter tuning and scale-out in three protocols for communicating model weights in asynchronous stochastic gradient descent.
We divide the learning rate by the average staleness of gradients, resulting in faster convergence and lower test error.
Our experiments show that the 1-\emph{softsync} protocol (in which the parameter server accumulates $\lambda$ gradients before updating the weights) minimizes gradient staleness and achieves the lowest runtime for a given test error.
We found that to maintain a model accuracy, it is necessary to reduce the \minibatch size as the number of learners is increased.
This suggests an upper limit on the level of parallelism that can be exploited for a given model, and consequently a need for algorithms that permit training over larger batch sizes.

\section*{Acknowledgement}

The work of Fei Wang is partially supported by National Science Foundation under Grant Number IIS-1650723.

\small
\bibliographystyle{abbrv}
\bibliography{sigproc}

\begin{thebibliography}{10}

\bibitem{bengio2012practical}
Y.~Bengio.
\newblock Practical recommendations for gradient-based training of deep
  architectures.
\newblock In {\em Neural Networks: Tricks of the Trade}, pages 437--478.
  Springer, 2012.

\bibitem{breuel2015effects}
T.~M. Breuel.
\newblock The effects of hyperparameters on {SGD} training of neural networks.
\newblock {\em arXiv:1508.02788}, 2015.

\bibitem{adam}
T.~Chilimbi, Y.~Suzue, J.~Apacible, and K.~Kalyanaraman.
\newblock Project {Adam}: Building an efficient and scalable deep learning
  training system.
\newblock OSDI'14, pages 571--582, 2014.

\bibitem{coates2013deep}
A.~Coates, B.~Huval, T.~Wang, D.~Wu, B.~Catanzaro, and N.~Andrew.
\newblock Deep learning with cots hpc systems.
\newblock In {\em Proceedings of the 30th ICML}, pages 1337--1345, 2013.

\bibitem{bounded-staleness}
H.~e.~a. Cui.
\newblock Exploiting bounded staleness to speed up big data analytics.
\newblock In {\em USENIX ATC'14}, pages 37--48, 2014.

\bibitem{distbelief}
J.~Dean, G.~S. Corrado, R.~Monga, K.~Chen, M.~Devin, Q.~V. Le, M.~Z. Mao,
  M.~Ranzato, A.~Senior, P.~Tucker, K.~Yang, and A.~Y. Ng.
\newblock Large scale distributed deep networks.
\newblock In {\em NIPS}, 2012.

\bibitem{duchi2011adaptive}
J.~Duchi, E.~Hazan, and Y.~Singer.
\newblock Adaptive subgradient methods for online learning and stochastic
  optimization.
\newblock {\em The Journal of Machine Learning Research}, 12:2121--2159, 2011.

\bibitem{ILSVRC15}
O.~R. et~al.
\newblock {ImageNet Large Scale Visual Recognition Challenge}.
\newblock {\em IJCV}, pages 1--42, 2015.

\bibitem{mpi}
M.~P.~I. Forum.
\newblock Mpi\ 3.0\ standard.
\newblock www.mpi-forum.org/docs/mpi-3.0/mpi30-report.pdf, 2012.

\bibitem{glorot2010understanding}
X.~Glorot and Y.~Bengio.
\newblock Understanding the difficulty of training deep feedforward neural
  networks.
\newblock In {\em AISTATS}, pages 249--256, 2010.

\bibitem{he2015delving}
K.~He, X.~Zhang, S.~Ren, and J.~Sun.
\newblock Delving deep into rectifiers: Surpassing human-level performance on
  imagenet classification.
\newblock {\em arXiv preprint arXiv:1502.01852}, 2015.

\bibitem{staleness}
Q.~Ho, J.~Cipar, H.~Cui, S.~Lee, J.~K. Kim, P.~B. Gibbons, G.~A. Gibson,
  G.~Ganger, and E.~P. Xing.
\newblock More effective distributed {ML} via a stale synchronous parallel
  parameter server.
\newblock In {\em NIPS 26}, pages 1223--1231. 2013.

\bibitem{ioffe2015batch}
S.~Ioffe and C.~Szegedy.
\newblock Batch normalization: Accelerating deep network training by reducing
  internal covariate shift.
\newblock {\em arXiv preprint arXiv:1502.03167}, 2015.

\bibitem{jia2014caffe}
Y.~Jia, E.~Shelhamer, J.~Donahue, S.~Karayev, J.~Long, R.~Girshick,
  S.~Guadarrama, and T.~Darrell.
\newblock Caffe: Convolutional architecture for fast feature embedding.
\newblock {\em arXiv preprint arXiv:1408.5093}, 2014.

\bibitem{krizhevsky2009learning}
A.~Krizhevsky and G.~Hinton.
\newblock Learning multiple layers of features from tiny images.
\newblock {\em Computer Science Department, University of Toronto, Tech. Rep},
  1(4):7, 2009.

\bibitem{krizhevsky2012imagenet}
A.~Krizhevsky, I.~Sutskever, and G.~E. Hinton.
\newblock Imagenet classification with deep convolutional neural networks.
\newblock In {\em Advances in neural information processing systems}, pages
  1097--1105, 2012.

\bibitem{vectorclock}
L.~Lamport.
\newblock Time, clocks, and the ordering of events in a distributed system.
\newblock {\em Commun. ACM}, 21(7):558--565, 1978.

\bibitem{lecun2015deep}
Y.~LeCun, Y.~Bengio, and G.~Hinton.
\newblock Deep learning.
\newblock {\em Nature}, 521(7553):436--444, 2015.

\bibitem{liu-asgd-nips-2015}
X.~{Lian}, Y.~{Huang}, Y.~{Li}, and J.~{Liu}.
\newblock {Asynchronous Parallel Stochastic Gradient for Nonconvex
  Optimization}.
\newblock {\em ArXiv e-prints}, June 2015.

\bibitem{logicaltime}
M.~Raynal and M.~Singhal.
\newblock Logical time: Capturing causality in distributed systems.
\newblock {\em Computer}, 29(2):49--56, 1996.

\bibitem{senior2013empirical}
A.~Senior, G.~Heigold, M.~Ranzato, and K.~Yang.
\newblock An empirical study of learning rates in deep neural networks for
  speech recognition.
\newblock In {\em ICASSP}, pages 6724--6728. IEEE, 2013.

\bibitem{parameterserver}
A.~Smola and S.~Narayanamurthy.
\newblock An architecture for parallel topic models.
\newblock {\em Proc. VLDB Endow.}, 3(1-2):703--710, 2010.

\bibitem{sutskever2013importance}
I.~Sutskever, J.~Martens, G.~Dahl, and G.~Hinton.
\newblock On the importance of initialization and momentum in deep learning.
\newblock In {\em Proceedinge of the 30th ICML}, pages 1139--1147, 2013.

\bibitem{zeiler2012adadelta}
M.~D. Zeiler.
\newblock Adadelta: An adaptive learning rate method.
\newblock {\em arXiv preprint arXiv:1212.5701}, 2012.

\bibitem{elastic-averaging-sgd}
S.~Zhang, A.~Choromanska, and Y.~LeCun.
\newblock Deep learning with elastic averaging {SGD}.
\newblock {\em CoRR}, abs/1412.6651, 2014.

\bibitem{mariana}
Y.~Zou, X.~Jin, Y.~Li, Z.~Guo, E.~Wang, and B.~Xiao.
\newblock Mariana: Tencent deep learning platform and its applications.
\newblock {\em Proc. VLDB Endow.}, 7(13):1772--1777, 2014.

\end{thebibliography}
\end{document}